# An efficient approach for finding the MPE in belief networks

**Zhaoyu Li**
Department of Computer Science
Oregon State University
Corvallis, OR 97331

**Bruce D'Ambrosio**
Department of Computer Science
Oregon State University
Corvallis, OR 97331

## Abstract

Given a belief network with evidence, the task of finding the $l$ most probable explanations (MPE) in the belief network is that of identifying and ordering the $l$ most probable instantiations of the non-evidence nodes of the belief network. Although many approaches have been proposed for solving this problem, most work only for restricted topologies (i.e., singly connected belief networks). In this paper, we will present a new approach for finding $l$ MPEs in an arbitrary belief network. First, we will present an algorithm for finding the MPE in a belief network. Then, we will present a linear time algorithm for finding the next MPE after finding the first MPE. And finally, we will discuss the problem of finding the MPE for a subset of variables of a belief network, and show that the problem can be efficiently solved by this approach.

## 1 Introduction

Finding the *Most Probable Explanation*(MPE) [21] of a set of evidence in a Bayesian (or belief) network is the identification of an instantiation or a *composite hypothesis* of all nodes except the observed nodes in the belief network, such that the instantiation has the largest posterior probability. Since the MPE provides the most probable states of a system, this technique can be applied to system analysis and diagnosis. Finding the $l$ most probable explanations of some given evidence is to identify the $l$ instantiations with the $l$ largest probabilities.

There have been some research efforts for finding MPE in recent years and several methods have been proposed for solving the problem. These previously developed methods can roughly be classified into two different groups. One group of methods consider the MPE as the problem of minimal-cost-proofs which works for finding the best explanation for text [11, 2, 31].
In finding the minimal-cost-proofs, a belief network is converted to Weighted Boolean Function Directed Acyclic Graphs (WBFDAG) [31], or cost-based abduction problems, and then the best-search techniques are applied to find MPE in the WBFDAGs. Since the number of the nodes in the converted graph is exponential in the size of the original belief network, efficiency of this technique seems not comparable with some algorithms directly evaluating belief networks [1]. An improvement is to translate the minimal-cost-proof problems into 0-1 programming problems, and solve them by using simplex combined with branch and bound techniques [24, 25, 1]. Although the new technique outperformed the best-first search technique, there are some limitations for using it, such as that the original belief networks should be small and their structures are close to and-or dags. The second group of methods directly evaluate belief networks for finding the MPE but restrict the type of belief networks to singly connected belief networks [21, 33, 34] or a particular type of belief networks such as BN2O [9], bipartite graphs [36]. Arbitrary multiply connected belief networks must be converted to singly connected networks and then are solved by these methods. The algorithm developed by J. Pearl [21] presents a message passing technique for finding two most probable explanations; but this technique is limited to only finding two explanations [17] and can not be applied to multiply connected belief networks. Based on the message passing technique, another algorithm [33, 34] has been developed for finding $l$ most probable explanations. Although this algorithm has some advantages over the previous one, it is also limited to singly connected belief networks.

In this paper, we will present an approach for finding the $l$ MPEs for arbitrary belief networks. First we will present an algorithm for finding the MPE. Then, we will present a linear time algorithm for finding the next MPE; so the $l$ MPEs can be efficiently found by activating the algorithm $l - 1$ times. Finally, we will discuss the problem of finding the MPE for a subset of variables in belief networks, and present an algorithm to solve this problem.

The rest of the paper is organized as follows. Section

2 present an algorithm for finding the MPE. Section 3 presents a linear time algorithm for finding the next MPE after finding the first MPE. Section 4 discusses the problem of finding the MPE for a subset of variables of a belief network. And finally, section 5 summarizes the research.

## 2 The algorithm for finding the MPE

There are two basic operations needed for finding the MPE: comparison for choosing proper instantiations and multiplication for calculating the value of the MPE. The difficulty of the problem of finding the MPE lies in finding or searching the right instantiations of all variables in a belief network since the multiplications for the MPE is simply given right instantiation of all variables. This means that finding the MPE can be a search problem. We can use search with back tracking techniques to find the MPE, but it may not be an efficient way because the search complexity is exponential with respect to the number of variables of a belief network in worst case.

We proposed a non-search method for finding the MPE. If we know the full joint probability of a belief network, we can obtain the $l$ MPEs by sorting the joint probability table in descending order and choosing the first $l$ instantiations. However, computing the full joint probability is quite inefficient. An improvement of the method is to use the "divide and conquer" technique. We can compute a joint probability distribution of some of distributions, find the largest instantiations of some variables in the distribution and eliminate those variables from the distribution; then, we combine the partially instantiated distribution with some other distributions until all distributions are combined together.

In a belief network, if a node has no descendants, we can find the largest instantiations of the node from its conditional distribution to support the MPE. In general, if some variables only appear in one distribution, we can obtain the largest instantiations of these variables to support the MPE. When a variable is instantiated in a distribution, the distribution is reduced and doesn't contain the variable; but each item of the reduced distribution is constrained by the instantiated value of that variable.

Given distributions of an arbitrary belief network, the algorithm for finding the MPE is:

1. For any node $x$ having no descendants, reduce its conditional distribution by choosing the largest instantiated values of the node for each instantiation of the other variables. The reduced distribution has no variable $x$ in it.

2. Create a factoring for combining all distributions;

3. Combine these distributions according to the factoring. If a result distribution of a conformal product (i.e. the product of two distributions) contains a variable $x$ which doesn't appear in any other distribution, reduce the result distribution (as in step 1), so that the reduced distribution doesn't contain variable $x$ in it.

The largest instantiated value of the last result distribution is the MPE[1].

Figure 1 is a simple belief network example to illustrate the algorithm. Given the belief network in figure 1, we want to compute its MPE. There are six distributions in the belief network. We use $D(x, y)$ to denote a distribution with variables $x$ and $y$ in it and $d(x = 1, y = 1)$ to denote one of items of the $D(x, y)$. In the step 1 of the algorithm, the distributions relevant to nodes $e$ and $f$ are reduced. For instance, $p(f|d)$ becomes $D(d)$:
$d(d = 0) = 0.7 \ with \ f = 1$;
$d(d = 1) = 0.8 \ with \ f = 0$.
In step 2 a factoring should be created for these distributions. For this example we assume the factoring is:

$$(((D(a)*D(a,c))*(D(b)*D(a,b,d)))*(D(c,d)*D(d))).$$

In step 3, these distributions are combined together some combined distributions are reduced if possible. The final combined distribution is $D(c, d)$:
$d(c = 1, d = 1) = .0224 \ with \ a = 1 \ b = 0 \ e = 1 \ f = 0$;
$d(c = 1, d = 0) = .0753 \ with \ a = 0 \ b = 0 \ e = 1 \ f = 1$;
$d(c = 0, d = 1) = .0403 \ with \ a = 0 \ b = 1 \ e = 1 \ f = 0$;
$d(c = 0, d = 0) = .1537 \ with \ a = 0 \ b = 0 \ e = 0 \ f = 1$.
Choosing the largest instantiation of $D(c, d)$, the MPE is: $p(a = 0, b = 0, c = 0, d = 0, e = 0, f = 1)$. If an unary operator $\Phi_x$ is defined for a probability distribution $p(y|x)$, $\Phi_x p(y|x)$, to indicate the operation of instantiating the variable $x$ and eliminating the variable from the distribution $p(y|x)$, the computations above for finding the MPE can be represented as:

$$\Phi_{c,d}(\Phi_a((p(a)*p(a,c))*\Phi_b(p(b)*p(a,b,d)))$$
$$*(\Phi_e p(e|c,d)*\Phi_f p(f|d))).$$

The most time consuming step in the algorithm is step 3. In step 1, the comparisons needed for instantiating a variable of a distribution is exponential in the number of conditioning variables of that variable. This cost is determined by the structure of a belief network. Factoring in step 2 could be arbitrary. In step 3, total computational cost consists of multiplications for combining distributions and comparisons for instantiating some variables in some intermediate result distributions. The number of variables of a conformal product or an intermediate result distribution is usually great than the that of distributions in step 1. If we use the *maximum dimensionality* to denote the maximum number of variables in conformal products, the time complexity of the algorithm is exponential with respect to the maximum dimensionality.

---

[1]Step 2 and step 3 can be mixed together by finding a partial factoring for some distributions and combining them.

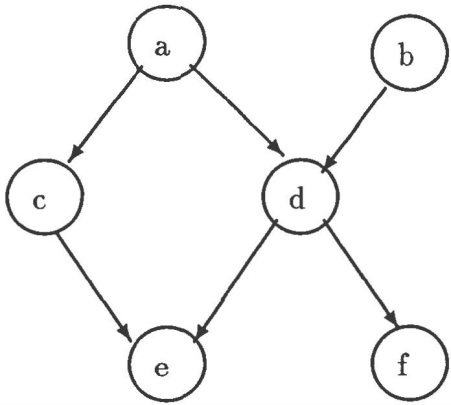

p(a): p(a=1)=0.2
p(b): p(b=1)=0.3
p(c|a): p(c=1|a=1)=0.8  p(c=1|a=0)=0.3
p(d|a,b): p(d=1|a=1,b=1)=0.7
p(d=1|a=1,b=0)=0.5
p(d=1|a=0,b=1)=0.5
p(d=1|a=0,b=0)=0.2
p(e|c,d): p(e=1|c=1,d=1)=0.5
p(e=1|c=1,d=0)=0.8
p(e=1|c=0,d=1)=0.6
p(e=1|c=0,d=0)=0.3
p(f|d): p(f=1|d=1)=0.2  p(f=1|d=0)=0.7

Figure 1: A simple belief network.

Step 2 is important to the efficiency of the algorithm because the factoring determines the maximum dimensionality of conformal products, namely the time complexity of the algorithm. Therefore, we consider the problem of efficiently finding the MPE as a factoring problem. We have formally defined an optimization problem, optimal factoring [16], for handling the factoring problem. We have presented an optimal factoring algorithm with linear time cost in the number of nodes of a belief network for singly connected belief networks, and an efficient heuristic factoring algorithm with polynomial time cost for multiply connected belief networks [16]. For reason of paper length, the optimal factoring problem will not be discussed here. The purpose of proposing the optimal factoring problem is that we want to apply some techniques developed in the field of combinatorial optimization to the optimal factoring problem, and apply the results from the optimal factoring problem to speedup the computation for finding the MPE.

It should be noticed that step 2 of the algorithm is a process of *symbolic reasoning*, having nothing to do with probability computation. There is a trade-off between the symbolic reasoning and probability computation. We want to use the polynomial time cost of this symbolic reasoning process to reduce the exponential time cost of the probability computation.

## 3  Finding the $l$ MPEs in belief networks

In this section, we will show that the algorithm presented in section 2 provides an efficient basis for finding the $l$ MPEs. We will present a linear time algorithm for finding next MPE. The $l$ MPEs can be obtained by first finding the MPE and then calling the linear algorithm $l-1$ times to obtain next $l-1$ MPEs.

### 3.1  Sources of the next MPE

Having found the first MPE, we know the instantiated value of each variable and the associated instantiations of the other variables in the distribution in which the variable was reduced. It is obvious that the instantiated value is the largest value of all instantiations of the variable with the same associated instantiations for the other variables in the distribution. If we replace that value with the second largest instantiation of the variable at the same associated instantiations of the other variables in the distribution, the result should be one of candidates for the second MPE. For example, if $d(a = A_1, b = B_1, ..., g = G_1)$ is the instantiated value for the first MPE when the variable $a$ is instantiated, the value $d(a = A_1, b = B_1, ..., g = G_1)$ is the largest instantiation of the variable $a$ with $b = B_1, ..., g = G_1$. If we replace $d(a = A_1, b = B_1, ..., g = G_1)$ with $d(a = A_2, b = B_1, ..., g = G_1)$, the second largest instantiation of $a$ given the same instantiation of $B$ through $G$, and re-evaluate all nodes on the path from that reduction operation to the root of the factor tree, the result is one of the candidates for the second MPE.

The total set of candidates for the second MPE comes from two sources. One is the second largest value of the last conformal product in finding the first MPE; and the other is the largest value of instantiations computed in the same computation procedure as for finding the first MPE but replacing the largest instantiation of each variable independently where it is reduced with the second largest instantiation. The similar idea can be applied for finding the third MPE, and so on.

The factoring (or the evaluation tree) generated in step 2 of the algorithm in section 2 provides a structure for computing those candidates. We use the example in that section to illustrate the process.

Figure 2 is the evaluation tree for finding the MPE for the belief network in figure 1 section 2. Leaf-nodes of the evaluation tree are the original probability distributions of the belief network. The meaning of an interior node is same as that we used in previous sections. The MPE is the $d(c=0, d=0)$ of the node $D(c,d)$ connecting to the root node, with instantiations $a=0$, $b=0$, $e=0$ and $f=1$. If we find the second largest $d(c=0, d=0)$ (with a different instantiation for variables a, b, e and f), to replace the largest $d(c=0, d=0)$ in $D(c,d)$, then the second MPE is the largest item in the revised $D(c,d)$. The second largest $d(c=0, d=0)$ comes from either by multiplying the largest value of $d(c=0, d=0)$ contributed from its left child node with the second largest value of $d(c=0, d=0)$ from its right child node, or by multiplying the largest value of $d(c=0, d=0)$ from its right child node with the second largest value of $d(c=0, d=0)$ from its left child node. The problem of finding the second largest $d(c=0, d=0)$, therefore, can be decomposed into the problem of finding the second largest $d(c=0, d=0)$ in each child node of the $D(c,d)$ node, and so on recursively.

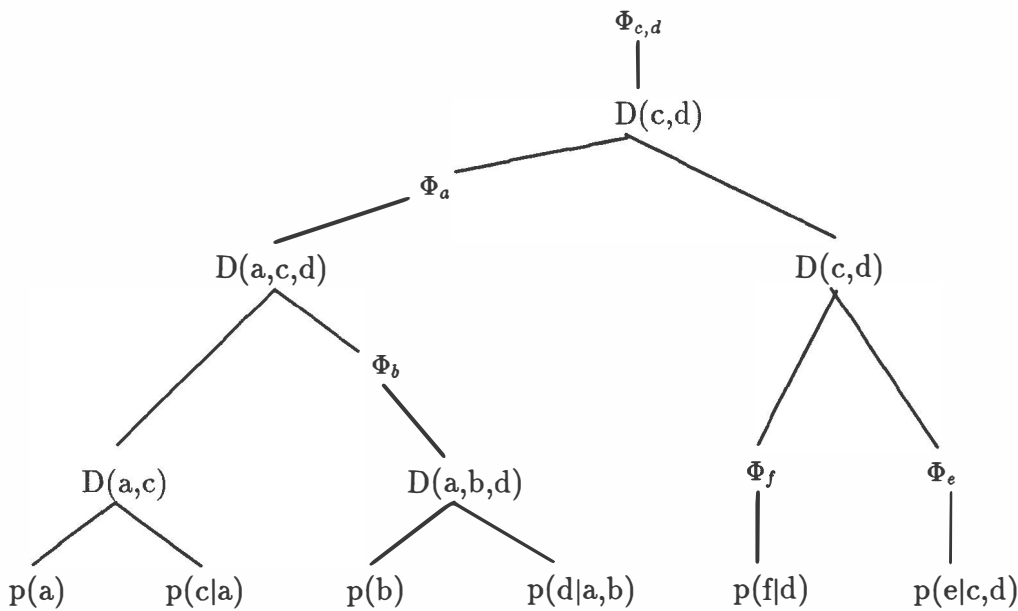

Figure 2: The evaluation tree for finding the MPE.

### 3.2 The algorithm for finding the next MPE

In order to efficiently search for the next MPE, we re-arrange the computation results from finding the first MPE. The re-arrangement produces a new evaluation tree from the original evaluation tree, so that a sub-tree rooted at a node meets all constraints (variable instantiations) from the root of the tree to that node.

**Evaluation Tree Re-arrangement** The rules for converting the original evaluation tree to the new evaluation tree are as follows. If a node is $\Phi_{x,y,...,z}$, duplicate the sub-tree rooted at the $\Phi$ node; the number of the sub-trees is equal to all possible instantiations of $\{x, y, ..., z\}$, and each sub-tree is constrained by one instantiation across $\{x, y, ..., z\}$. If a node is a conformal product node, nothing needs to be done. If a node has no $\Phi$ nodes in its sub-tree, prune the node and its sub-tree because all probabilistic information about the node and its sub-tree are known at its parent node. Figure 3 is an evaluation tree generated from the evaluation tree in figure 2. The evaluation tree in figure 3 is not complete; we only draw one branch of each $\Phi$ node.

**Marking the Evaluation Tree** The evaluation tree is annotated with marks to indicate the MPE's that have been returned. In figure 3 these marks are contained as the arguments to the *max* annotation at each node. There are two meanings for the parameters of *max*, depending on whether it is attached to a $\Phi$ or conformal-product node. An integer at a node denotes the ranking of the corresponding instantiated value contributed from its child node. For example, the first 1 at the root node indicates that the node contains the largest value of $d(c=0, d=0)$, and the "*" indicates that the value was used in a previous MPE (the first, in this case). The second 1 carries corresponding information for $d(c=1, d=0)$. For the conformal product immediately below the root node, the first 1 indicates the largest value of $d(c=0, d=0)$ has been retrieved from its left child node and the right 1 indicates the largest value of $d(c=0, d=0)$ has been retrieved its right child node.

**The Max Method** The *max* method on an evaluation tree is defined as follows:

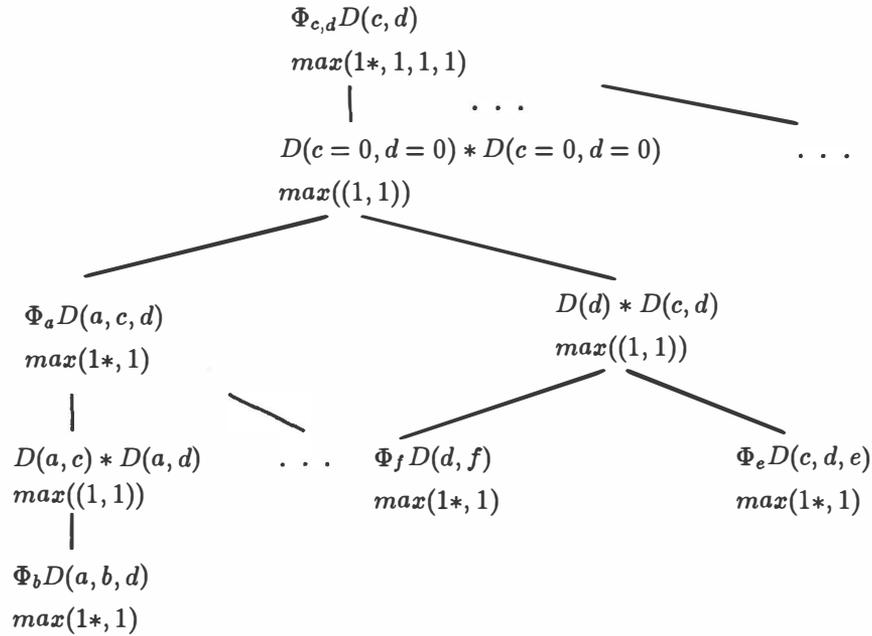

Figure 3: The evaluation tree for finding the next MPE.

1. If a parameter is marked, i.e. its corresponding instantiated value was used for finding the previous MPE, generate the next instantiation: query ($max$) its child nodes to find and return the instantiated values matching the ranking parameters (we will discuss the determination of the parameters later).

2. If no parameter is marked, mark one parameter which corresponds to the largest instantiated value of the node, and return the value to its parent node.

**The Gen Method** We define a method $gen$ to generate next ranking parameter for an integer i: $gen(i) = i+1$ if (i+1) is in the domain, otherwise $gen(i) = 0$. The $gen$ method for generating next possible ranking pairs of integers can be defined as follows. If current ranking pair is $(i,j)$, then the next possible ranking pairs are generated:

1. If $(i-1, j+1)$ exists then $gen(i,j) = (i, j+1)$;
2. If $(i+1, j-1)$ exists then $gen(i,j) = (i+1, j)$.

The pairs $(0, x)$ and $(x, 0)$ exist by definition when $x$ is in a valid domain size; $gen$ will generate $(1, x+1)$ and $(x+1, 1)$ when applied to $(1, x)$ and $(x, 1)$. The range of an integer in a node is from 1 to the product of the domain size of these variables of $\Phi$ nodes in the sub-tree of that node. A pair of integer is valid if each integer in it is in the range.

Given the evaluation tree and the defined methods $max$ and $gen$ for each node, the procedure for finding the next $l$ MPEs is: activate the $max$ method of the root node $l$ times.

### 3.3 Analysis of the algorithm

The algorithm described above returns the next MPE every time it is called from the second MPE. First, we will show that the algorithm is complete; that is, it can find every possible instantiation of variables in a belief network. According to the rules for creating an evaluation tree, the number of different paths from the root to all leaves in the evaluation tree is equal to the product of domain size of all variables in the belief network. That is, each path corresponds to an instantiation. Since the $max$ method will mark each path it has retrieved during finding each successive MPE, and will not retrieve a marked path, the algorithm retrieves each path exactly once.

Second, the algorithm will always find the next MPE. When querying for the next MPE, the root node of the evaluation tree is queried to find a candidate which has the same instantiation for the variables in the root node as that for the previously found MPE, but has next largest value. This computation is decomposed into the same sub-problems and passed to its child nodes, and from its child nodes to their child nodes, and so on. Each node being queried will return next largest value to its parent node or will return 0 if no value can be found. Returning next largest value from a node to its parent node is ensured by the $gen$ and $max$ methods. The $gen$ method determines which instantiated value should be obtained from its child nodes. If the $gen$ method has one integer as parameter, it generates the successor of the integer or a zero as we expected. If the *gen* has a pair of integers as its parameter, we know, from the definition of the *gen* method, that the pair $(i, j + 1)$ is generated only if $(i - 1, j + 1)$ exists; the pair $(i + 1, j)$ is generated only if $(i + 1, j - 1)$ exists. On the other hand, if $(i, i)$ is marked, it will not generate $(i, i+1)$ or $(i+1, i)$ unless $(i - 1, i)$ or $(i, i - 1)$ exist. Therefore, *gen* only generates the pair needed for finding next largest value in a node. Choosing the largest value from a list of instantiated values in *max* is obvious. From this we can conclude that the algorithm will always retrieve the next MPE each time it is called.

The time complexity of the algorithm for finding the next MPE in a belief network is linear in the number of instantiated variables in the evaluation tree. At a $\Phi$ node, only one marked value must be replaced by a new value. Therefore, only one child node of a $\Phi$ node needs exploring. AT a conformal product node, there is at most one value to be requested from each child node according to the definition of *gen*. So, each child node of a conformal product node will be explored at most once. For example, after $gen(1, 2)$ generates $(1, 3)$, and $gen(2, 1)$ generates $(2, 2)$ and $(3, 1)$, when $(2, 2)$ is chosen, there is no query for $(2, 2)$ because the instantiated values for $(2, 2)$ can be obtained from $(1, 2)$ and $(2, 1)$ of previous computation. Therefore there are at most $n$ $\Phi$ nodes plus $(n-1)$ conformal product nodes in an evaluation tree to be visited for finding next MPE, where $n$ is the number of nodes in the belief network. Also there is a *max* operation in each node of the evaluation tree and only one or two multiplications needed in a conformal product node. Therefore, the algorithm for finding the next MPE is efficient.

The time complexity for converting a factoring to the evaluation tree for finding next MPE should be no more than that for computing the first MPE. This conversion is the process of data rearrangement which can be carried out simultaneously with the process for finding the first MPE.

The space complexity of the algorithm is equal to the time complexity for finding the first MPE, since this algorithm saves all the intermediate computation results for finding next MPE. The time complexity for finding the MPE in a singly connected belief network is $O(k * 2^n)$, where $k$ is the number of non-marginal nodes of the belief network and $n$ is the largest size of a node plus its parents in the belief network. Considering that the input size of the problem is in the order of $O(2^n)$, the space complexity is at most $k$ times of the input size for singly connected belief networks. For a multiply connected belief network, the time complexity for finding the MPE can be measured by the maximum dimensionality of conformal products, which is determined by both the structure of a belief network and the factoring algorithm. The time complexity for finding the MPE in terms of input is exponential with respect to the difference between the maximum dimensionality for finding the MPE and the largest size of a node plus its parent nodes in the belief network. This time complexity reflects the hardness of the belief network if the factoring for it is optimal. If the factoring is optimal, the time and space complexity are the best that can be achieved for finding the $l$ MPEs.

## 4 The MPE for a subset of variables in belief networks

In this section, we will discuss the problem of finding the MPE for a subset of variables in belief networks. We will show that finding the MPE for a subset of variables in a belief network is similar to the problem of finding the MPE over all variables in the belief network, and the problem can be considered as an optimal factoring problem. Therefore, the algorithm for finding the MPE for a subset of variables in a belief network, either singly connected or multiply connected, can be obtained from the algorithm in section 2 with little modifications.

We first examine the differences between probabilistic inference (posterior probability computation) and finding the MPE for all variables in a belief network so that we can apply the approach described in section 2 to the problem of finding the $l$ MPEs for a subset of variables. There are three differences. First, there is a target or a set of queried variables in posterior probability computation; but there is no target variable in finding the MPE. The computation for a posterior probability computation is query related and only the nodes relevant to the query are involved in the computation, whereas finding the MPE relates to whole belief network. Second, the addition operation in summing over variables in posterior probability computation are replaced by comparison operation in finding the MPE, but the number of operations in both cases is the same. And finally, variables with no direct descendants in a distribution can be reduced at the beginning of finding the MPE whereas queried variables cannot be summed over in posterior probability computation.

Finding the MPE for a set of variables in belief networks combines elements of the procedures for find the MPE and for posterior probability computation. Since not all variables in a belief network are involved in the problem of finding the MPE for a set of variables, the variables not relevant to the problem can be eliminated from computation. Therefore, two things should be considered in finding the MPE for a set of variables in a belief network. One thing is to choose relevant nodes or distributions for computation. The second is to determine the situation in which a variable can be summed over or reduced. The first is simple because we can find the relevant nodes to some queried nodes given some observed nodes in linear time with respect to the number of nodes in a belief network[6, 29]. We have the following lemmas for determining when a node can be summed over or reduced.

Suppose we have the variables relevant to a set of queried variables for finding the MPE given some observations. These variables can be divided into two sets: a set $\Phi$ which contains the queried variables (or the target variables for finding the MPE) and a set $\Sigma$ which contains the rest of variables (or variables to be summed over in computation). The current distributions are represented by $D_i$ for $1 \leq i \leq n$ and the variables in a distribution $D_j$ are also represented in the set $D_j$.

**Lemma 1** *Given $\alpha \in \Sigma$, if $\alpha \in D_i$ and $\alpha \notin D_j$ for $i \neq j$, $1 \leq j \leq n$, then $\alpha$ can be summed over from the distribution $D_i$.*

Proof: The lemma is obvious. It is the same situation in which we sum over some variables in posterior probability computation. □

**Lemma 2** *Given $\alpha \in \Phi$, if $\alpha \in D_i$ and $\alpha \notin D_j$ for $i \neq j$, $1 \leq j \leq n$, and for any other $\beta \in D_i$, $\beta \in \Phi$, then distribution $D_i$ can be reduced with respect to $\alpha$.*

Proof: Since $\alpha \in \Phi$ and $\alpha \in D_i$ only, the information relevant to $\alpha$ is in the distribution $D_i$. So, we can instantiate variable $\alpha$ to find its largest instantiated value to contribute the MPE, and the reduced distribution of $D_i$ contains all possible combinations cross values of other variables in $D_i$. Since for any other $\beta \in D_i$, $\beta \in \Phi$, no summation for some other variables of $D_i$ afterward will affect the $\beta$. So $\beta$ can be instantiated later if possible. □

Given the two lemmas, the algorithm in section 2 can be modified for finding the MPE for a subset of variables in belief networks. Given a belief network, a set of variables $\Phi$ and evidence variables $E$, the algorithm for finding the MPE of $\Phi$ is:

1. Find variables of $T$ which are the predecessors of variables in set $\Phi$ or $E$ and connected to set $\Phi^2$. The distributions relevant to the variables in $T$ are needed for finding the MPE of $\Phi$.

2. For any variable $x$ of $T$ having no descendants in the belief network, reduce the conditional distribution of the node $x$ by choosing the items of the distribution which have the largest instantiated values of $x$ with same associated instantiations for the other variables. The reduced distribution has no variable $x$ in it.

3. Create a factoring for all distributions;

4. Combine these distributions according to the factoring. Apply lemma 1 and lemma 2 to each result distribution in probability computation. If both lemmas apply to a distribution, apply lemma 1 first.

Take the belief network in figure 1 as an example. We want to find the MPE for the variables $\Phi = \{c, d, e\}$

---
[2] An evidence node breaks the connection of the node with its child nodes.

given $E$ is empty. In the step 1 of the algorithm, the variables related to the query are found, $T = \{a, b, c, d, e\}$. In the step 2, distribution $D(c, d, e)$ is reduced to $D(c, d)$. In the step 3, assume a proper factoring is found:

$$((D(a) * D(a,c)) * (D(b) * D(a,b,d))) * D(c,d).$$

In step 4, combine these distributions according to the above factoring and apply lemma 1 or/and lemma 2 to any result distribution if applicable. Then we obtain the MPE for variables $\{c, d, e\}$. The whole computation can be represented as:

$$\Phi_{c,d}(\sum_a((p(a)*p(c|a))*(\sum_b(p(b)*p(d|a,b))))*\Phi_e p(e|c,d)).$$

This algorithm is very similar to the algorithm in section 2. Since the time complexity of the first step of the algorithm is linear with respect to the number of variables in belief networks, the most time consuming step of the algorithm is the step 4 which is determined by the factoring result of the step 2. Therefor, efficiently finding the MPE for a set of variables in a belief network can be considered as an optimal factoring problem. By using the algorithm presented in the previous section after finding the first MPE, the problem of finding the $l$ MPEs for a set of variables can be easily solved.

In this section we have presented an algorithm for the problem of finding the MPE for a set of variables in a belief network and shown that the problem can be efficiently solved through an optimal factoring problem. However, we don't present a factoring algorithm for this problem here. We have discussed the difference between this problem and the problem of finding the MPE for all variables in a belief network, and the difference between this problem and the problem of computing posterior probability of a set of variables. So, we can apply the factoring strategies developed for posterior probability computation or for finding the MPE for whole belief network to this problem. It might be that a more efficient factoring algorithm exists for this problem. However, we will not discuss this further or present any algorithm for the problem in this paper.

## 5 Related work

Dawid [5] pointed out that the problem of finding the MPE of a belief network can be simply realized by replacing the normal marginalization operation of the distribution phase of evidence propagation in a join-tree in posterior probability computation by max-marginalization (i.e. taking max instead of summing). Therefore, the efficiency of an algorithm for finding the MPE depends basically on the corresponding posterior probability computation algorithm. Golmard developed an algorithm for finding the MPE independent of our work [7]. We have requested a copy of the work and are waiting to receive it.

## 6  Conclusion

In this paper we have presented a framework, optimal factoring, for finding the most probable explanations (MPE) in a belief network. Under this framework, efficiently finding the MPE can be considered as the problem of finding an ordering of distributions in the belief network and efficiently combining them. The optimal factoring framework provides us many advantages for solving the MPE problem. First, the framework reveals the relationship between the problem of finding the MPE and the problem of querying posterior probability. Second, quantitative description of the framework provides a way of measuring and designing an algorithm for solving the problem. Third, the framework can be applied to both singly connected belief networks and multiply connected belief networks. Forth, the framework can be applied to the problem of finding the MPE for a set of variables in belief networks. Finally, the framework provides a linear time algorithm for finding next MPE. Under the optimal factoring framework, We have developed an optimal factoring algorithm for finding the MPE for a singly connected belief network. We have also developed an efficient algorithm for finding the MPE in multiply connected belief networks.